\documentclass[conference]{IEEEtran}
\IEEEoverridecommandlockouts
\usepackage{cite}
\usepackage{amsmath,amssymb,amsfonts}
\usepackage{algorithmic}
\usepackage{graphicx}
\usepackage{textcomp}
\usepackage{xcolor}
\usepackage{url}
\usepackage{booktabs}
\usepackage{pdfpages} 

\def\BibTeX{{\rm B\kern-.05em{\sc i\kern-.025em b}\kern-.08em
    T\kern-.1667em\lower.7ex\hbox{E}\kern-.125emX}}

\begin{document}

\title{A Decentralized LiDAR-SLAM System with Certifiably Optimal Pose Graph Optimization}

\author{\IEEEauthorblockN{Baoshan Song}
\IEEEauthorblockA{\textit{Department of Aeronautical and}\\{\textit{Aviation Engineering}}\\
\textit{The Hong Kong Polytechnic University}\\
Hong Kong, China \\
baoshan.song@connect.polyu.hk}
\and
\IEEEauthorblockN{Feng Huang}
\IEEEauthorblockA{\textit{Department of Aeronautical and}\\{\textit{Aviation Engineering}}\\
\textit{The Hong Kong Polytechnic University}\\
Hong Kong, China \\
darren.huang@polyu.edu.hk}
\and
\IEEEauthorblockN{Li-Ta Hsu\textsuperscript{*}}
\IEEEauthorblockA{\textit{Department of Aeronautical and}\\{\textit{Aviation Engineering}}\\
\textit{The Hong Kong Polytechnic University}\\
Hong Kong, China \\
lt.hsu@polyu.edu.hk}
}

\maketitle

\begin{abstract}

Decentralized multi-robot LiDAR-SLAM is essential for collaborative missions but faces significant challenges in maintaining global consistency. Existing frameworks predominantly rely on local-search optimization or one-time coordinate alignment, which are prone to suboptimal convergence and long-term inconsistency, especially in large-scale or degenerate environments. To address these limitations, this paper presents the first decentralized LiDAR-SLAM system that integrates a state-of-the-art certifiably optimal Pose Graph Optimization (PGO) backend. By leveraging the Riemannian Block Coordinate Descent (RBCD) algorithm, our system ensures globally consistent trajectory estimation without requiring accurate initial guesses. Experimental results demonstrate that the proposed framework achieves superior robustness, improving trajectory RMSE by up to 48.9\% compared to the state-of-the-art DiSCo-SLAM.
\end{abstract}

\section{Introduction}
Decentralized multi-robot LiDAR-SLAM is essential for GNSS-denied environments \cite{suzuki_gnss_2022,liu_glio_2023,zhang_gnss_2024,huang_roadside_2025,guadagnino_kiss-slam_2025}, yet inherent drift in weakly observable directions poses a significant challenge. While existing frameworks prioritize front-end speed \cite{zhu_swarm-lio2_2024,kim_skidslam_2025} or simplify back-end registration \cite{huang_disco-slam_2021}, they often rely on local search-based PGO \cite{zhong_dcl-slam_2024} that is prone to local minima and lacks optimality guarantees. To bridge the gap between efficiency and accuracy, this paper integrates the Riemannian Block Coordinate Descent (RBCD) algorithm \cite{tian_distributed_2021} into a decentralized LiDAR-SLAM pipeline. By formulating PGO as a semidefinite programming (SDP) problem solved on Riemannian manifolds, our system ensures computational speed while providing certifiably optimal certificates. Real-world experiments confirm its superior trajectory consistency over mainstream local estimators.
\begin{figure}
    \centering
    \includegraphics[width=1\linewidth]{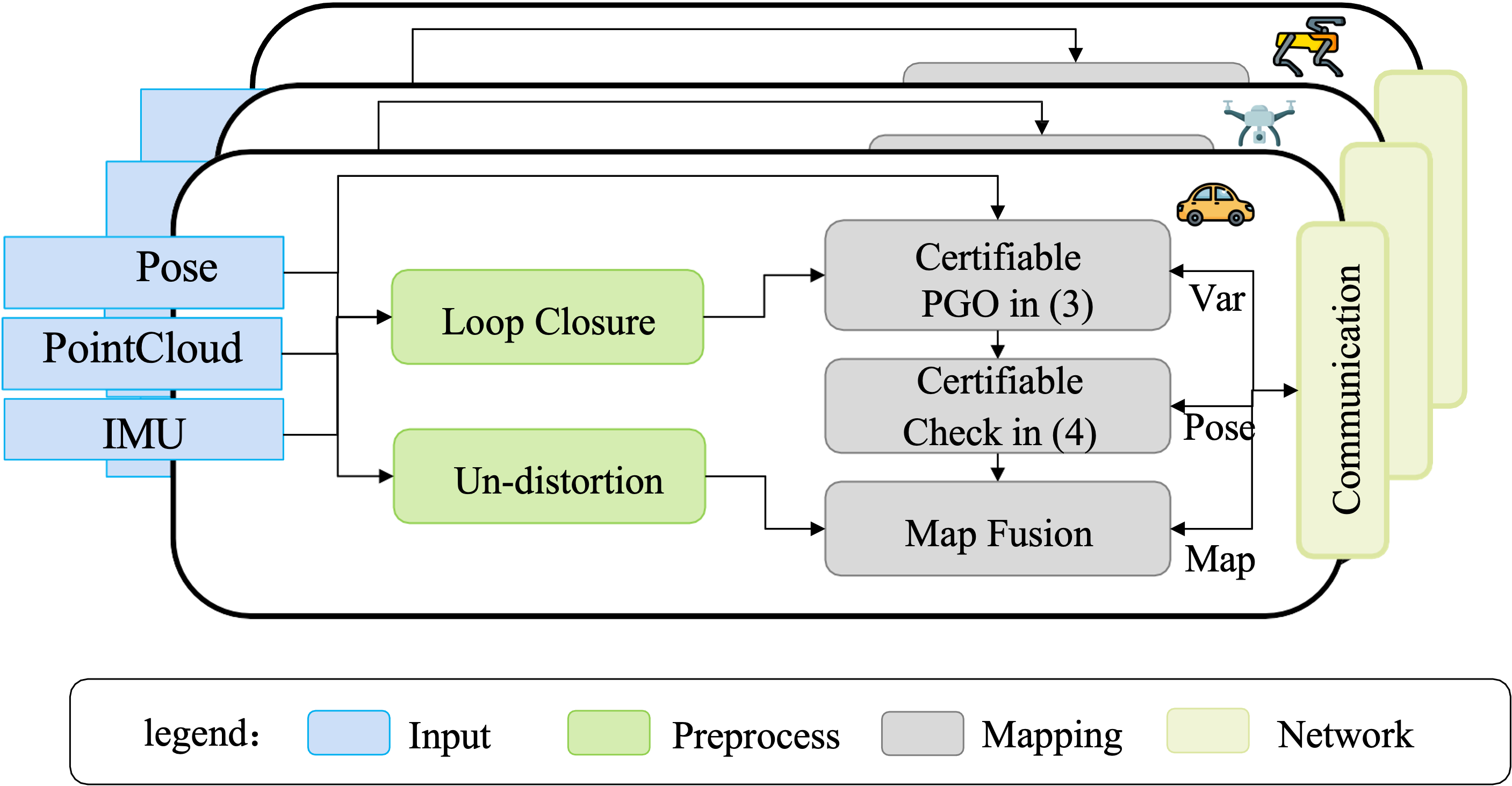}
    \caption{Pipeline of the proposed system. The main contribution compared to DiSCo-SLAM is the certifiably optimal back-end PGO.}
    \label{fig:pipeline}
\end{figure}

\section{Methodology}
\subsection{System Architecture and Overview}
The proposed decentralized multi-robot LiDAR-SLAM system is designed as a modular peer-to-peer (P2P) framework to ensure scalability and robustness in bandwidth-constrained environments. As illustrated in Fig. \ref{fig:pipeline}, the front-end pipeline adopts point clouds and odometry estimates from LIO-SAM\cite{shan_lio-sam_2020} as inputs. Note that the inertial measurement units (IMU) measurements are only used for point cloud un-distortion. Following the decentralized architecture established in DiSCo-SLAM~\cite{huang_disco-slam_2021}, our system utilizes Scan-Context descriptors~\cite{kim_scan_2018} for global loop closure detection and an asynchronous P2P communication protocol for sub-map synchronization. These modules collectively provide the inter-robot geometric constraints required for our core contribution: a certifiably optimal backend. Unlike the local-search optimization used in prior works, we utilize RBCD\cite{tian_distributed_2021} to recover certified global trajectories, which are subsequently used for consistent TSDF-based volumetric map fusion\cite{dubois_dense_2020}.

\subsection{Decentralized Robot Alignment via Certifiable Optimization}

\subsubsection{Limitations of Local Searching based PGO}
Prior decentralized systems, such as DiSCo-SLAM\cite{huang_disco-slam_2021}, typically address the inter-robot initialization problem by solving for a one-time rigid body transform $T_{ab} \in \mathrm{SE}(3)$ upon the first encounter:
\begin{equation}
    T_{ab} = \arg\min_{T \in \mathrm{SE}(3)} \left\| T \hat{T}_a  \hat{T}_b^{-1} \right\|^2
\end{equation}
where $\hat{T}_a$ and $\hat{T}_b$ are matched keyframe poses at the rendezvous moment. This formulation possesses two critical structural limitations that impede long-term consistency: 1) The global consistency of the multi-robot map relies entirely on a single registration event, which is prone to failure in geometrically degenerate environments (e.g., long corridors or featureless tunnels). 2) Subsequent inter-robot loop closures encountered during the mission are not utilized to refine the initial $T_{ab}$. Consequently, any odometric drift accumulated after the first encounter remains uncorrected at the global level.

\subsubsection{Certifiable PGO with Lagrangian duality}
We overcome these limitations by treating all intra- and inter-robot constraints within a unified, continuous global PGO framework. Assuming there are $n$ unknown poses, the full pose graph $\mathcal{G} = (\mathcal{V}, \mathcal{E}_{\text{intra}} \cup \mathcal{E}_{\text{inter}})$ where $\mathcal{V}=[n]$ denotes the nodes and $\mathcal{E}=\mathcal{V}\times\mathcal{V}$ denotes the edges is optimized by minimizing the following objective:
\begin{equation}
    \min_{\{R_i, t_i\}} \sum_{(i,j) \in \mathcal{E}} \left( \kappa_{ij}\left\| R_j - R_i \tilde{R}_{ij} \right\|_F^2 + \sigma_{ij}\left\| t_j - t_i - R_i \tilde{t}_{ij} \right\|^2 \right)
\end{equation}
where $i,j\in[n]$ denotes the pose node index; $\kappa_{ij}$ and $\sigma_{ij}$ represent the information weights for rotation and translation. Unlike one-time alignment, this formulation allows every new rendezvous to propagate corrections throughout the entire joint trajectory.

This non-convex problem is solved using the RBCD algorithm\cite{tian_distributed_2021}, which operates on the Stiefel manifold and is equivalent to solving the semidefinite relaxation:
\begin{equation}
    \min_{Q \in \mathcal{S}^{dn}}\ \langle \mathcal{L}_\rho, Q \rangle \quad \text{subject to} \quad Q \succeq 0, \text{diag}(Q) = \mathbf{I}
\end{equation}
where $\mathcal{L}_\rho$ is the connection Laplacian assembled from all multi-robot edges. At convergence, RBCD provides a dual certificate matrix 
\begin{equation}
    S^* = \mathcal{L}_\rho - \mathrm{blockdiag}(\Lambda^*)
\end{equation}
Global optimality is verified by checking the positive semidefiniteness of the dual certificate matrix, $S^* \succeq 0$. Specifically, the relaxation is certified to be tight when the $(d+1)$-th smallest eigenvalue satisfies $\lambda_{d+1}(S^*) > 0$ (where $d=3$ for $SE(3)$), ensuring that the recovered solution is the unique global optimum. By leveraging this dual certificate, our system can explicitly detect and reject locally optimal solutions arising from degenerate inter-robot observations, providing the first structural integrity guarantee for decentralized LiDAR-SLAM.

\begin{figure}
    \centering
    \includegraphics[width=1\linewidth]{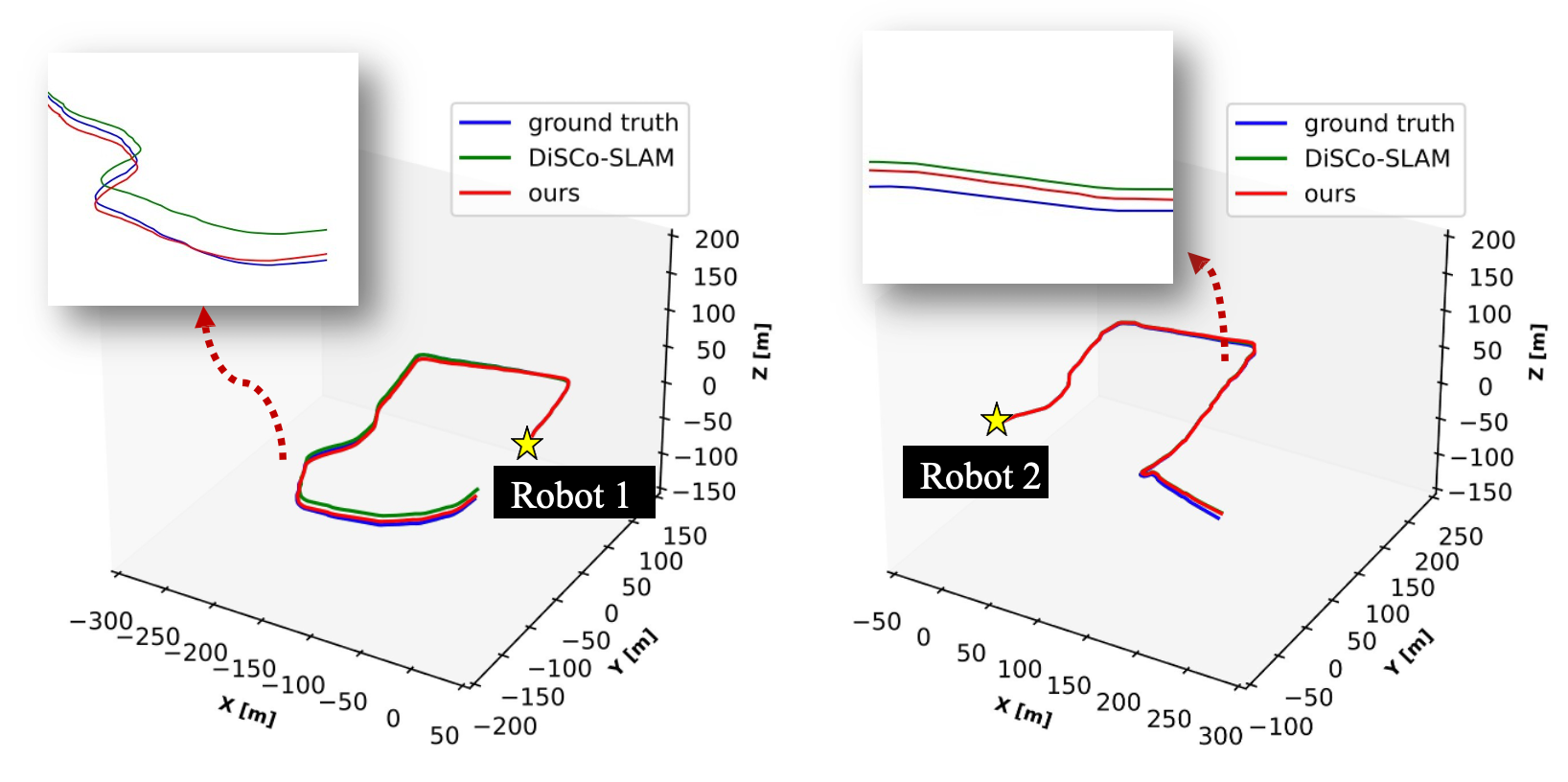}
    \caption{Trajectories of the compared methods}
    \label{fig:localization}
\end{figure}

\begin{figure}
    \centering
    \includegraphics[width=1\linewidth]{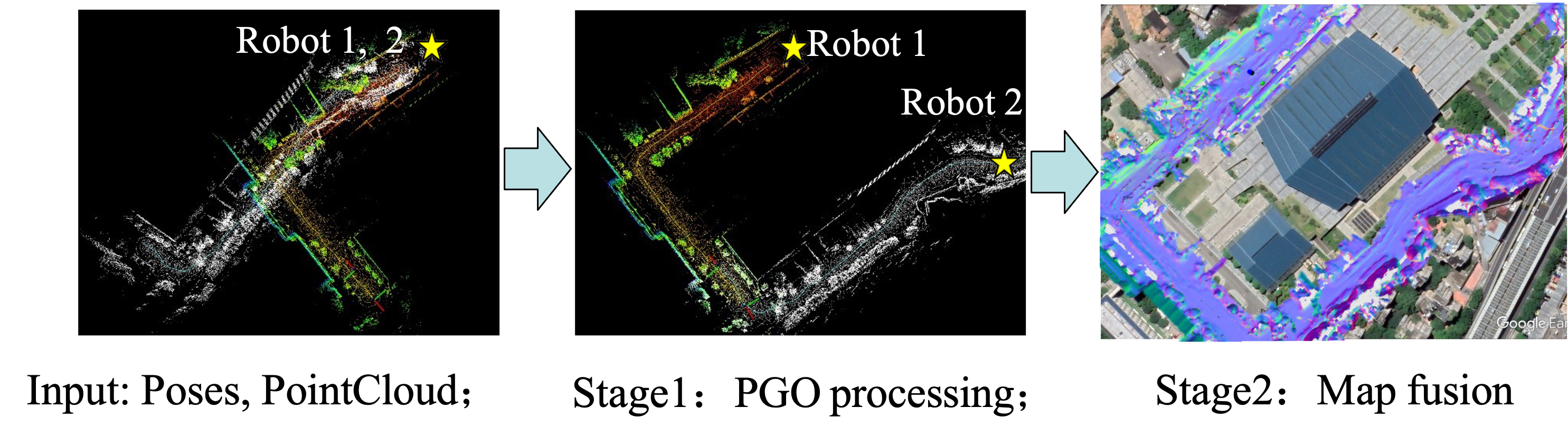}
    \caption{Process of the proposed decentralized LiDAR SLAM system}
    \label{fig:mapping}
\end{figure}

\section{Experimental Results and Analysis}

To evaluate the performance, we utilized a single vehicle to collect two sequential trajectories with overlapping regions, which were then treated as data from two independent robotic agents to simulate a decentralized multi-robot mission.
Our hardware platform comprises a 64-line Ouster LiDAR\cite{ouster_os0_2024} (downsampled to 32 lines), an ADI-16470 IMU\cite{analog_adis16470_2020}, and a ground-truth system using a tactical-grade GNSS/IMU system. We utilize LIO-SAM\cite{shan_lio-sam_2020} without loop closure as the base odometry (LIO-SAM-odometry). The proposed system is compared against DiSCo-SLAM, which is a distributed LiDAR-inertial SLAM system also employing the Scan-Context descriptor.
Localization results in Fig. \ref{fig:localization} and Table \ref{tab:rmse} demonstrate that our system effectively suppresses this positioning drifting. Compared to DiSCo-SLAM, our method improves the Root Mean Square Error (RMSE) for Robot 1 and Robot 2 by 48.9\% and 13.4\%, respectively. Compared to the raw LIO-SAM-odometry, the improvements reach 48.4\% and 51.4\%. The results show that while DiSCo-SLAM trajectories gradually deviate from the ground truth after long-term traveling, our approach maintains high-precision localization by providing globally optimal certificates for the pose graph. Finally, the mapping result is illustrated in Fig. \ref{fig:mapping}, which aligns accurately with the satellite map.

\begin{table}[h]
\caption{Trajectory Position RMSE (m) Comparison}
\label{tab:rmse}
\centering
\begin{tabular}{lccc}
\hline
\textbf{Robot}& \textbf{LIO-SAM-odometry}  & \textbf{DiSCo-SLAM} & \textbf{Ours} \\ \hline
Robot 1         & 7.01          & 7.09                     & \textbf{3.62   }             \\
Robot 2             & 4.90      & 2.73                 & \textbf{2.38 }                   \\ \hline
\end{tabular}
\end{table}

\section{Conclusion}
This paper presented the first decentralized LiDAR-SLAM system integrating a certifiably optimal PGO backend, effectively overcoming the local-minima and localization drifting in long-term operation in traditional methods. Experimental results in challenging environments confirm a significant 48.9\% accuracy improvement over DiSCo-SLAM, proving the strength of certifiable batch PGO for decentralized SLAM.
Our future works will focus on extending the system to support heterogeneous sensor suites, such as integrating visual-inertial modules to enhance robustness in feature-poor areas.

\bibliography{icra}
\bibliographystyle{IEEEtran}

\clearpage
\includepdf[pages=-, fitpaper=false]{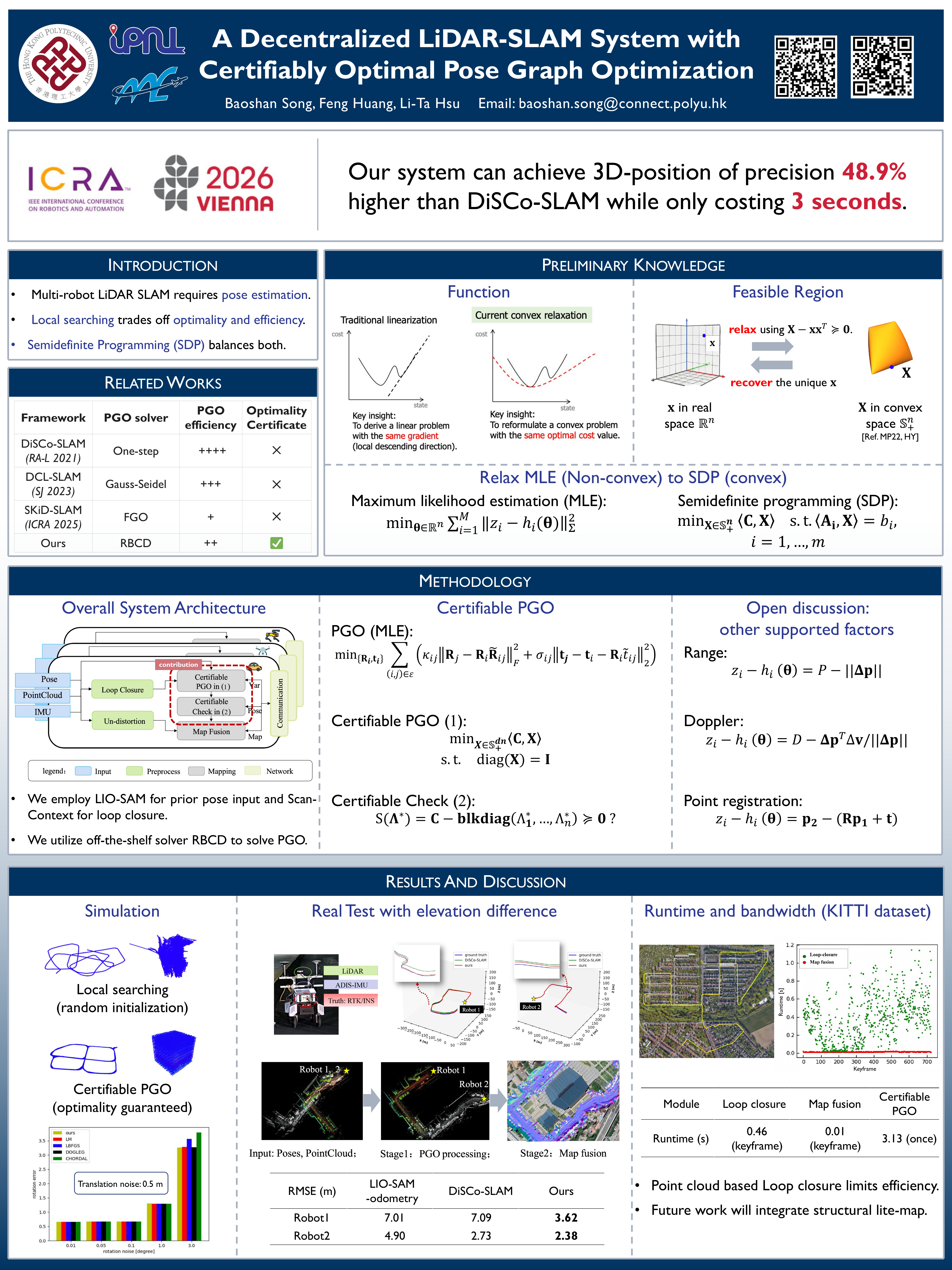}

\end{document}